# Lifted Relational Variational Inference

**Jaesik Choi** and **Eyal Amir**
Department of Computer Science
University of Illinois at Urbana-Champaign
Urbana, IL 61874, USA
{jaesik,eyal}@illinois.edu

## Abstract

Hybrid continuous-discrete models naturally represent many real-world applications in robotics, finance, and environmental engineering. Inference with large-scale models is challenging because relational structures deteriorate rapidly during inference with observations. The main contribution of this paper is an efficient relational variational inference algorithm that factors large-scale probability models into simpler variational models, composed of mixtures of iid (Bernoulli) random variables. The algorithm takes probability relational models of large-scale hybrid systems and converts them to a close-to-optimal variational models. Then, it efficiently calculates marginal probabilities on the variational models by using a latent (or lifted) variable elimination or a lifted stochastic sampling. This inference is unique because it maintains the relational structure upon individual observations and during inference steps.

## 1 Introduction

Many real-world systems can be described using continuous and discrete variables with relations among them. Such examples include measurements in environmental sensor networks, localizations in robotics, and economic forecasting in finance. In such large systems, efficient and precise inference is essential. As an example from environmental engineering, an inference algorithm can predict a posterior of unobserved groundwater levels and contamination levels at different locations, and making such an inference precisely is critical to decision makers.

Real-world systems have large numbers of variables including both discrete and continuous. Probabilistic first order languages, e.g. [3; 15; 24; 25; 14; 26; 28], describe probability distributions at a relational level with the purpose of capturing the structure of larger models. A key challenge of inference procedures with the languages is that they often result in intermediate density functions involving many random variables and complex relationship among them.

Lifted inference presently can address discrete models and continuous models, but not hybrid ones. For ($d$-valued) discrete variables, lifted inference can take an advantage of the insight which groups equivalent models into histogram representations with an order of $poly(d)$ entries [12; 22; 17] (instead of $exp(d)$ entries in traditional *ground* models). For Gaussian potentials, lifted inference can use an insight which enables maintaining compact covariance matrices during (and after) inference, e.g. [5; 7; 1].

Nonparametric variational models, e.g. NP-BLOG [4] and Latent Tree Models [32; 8], handle inference problems for discrete models and continuous Gaussian models. Here, our model and algorithms provide a solution for general (non-Gaussian) hybrid models.

In this paper, we first define **Relational Hybrid Models** (Section 2) and variational models (Section 3). We presents pragmatic algorithms (Section 4, 5 and 6) based on a new insight, **relational variational-inference lemmas** (Section 7), which accurately factors densities of relational models into mixtures of iid random variables. These lemmas enable us to build a variational approximation algorithm, which takes large-scale graphical models with hybrid variables and finds close to optimal relational variational models. Then, our inference algorithms, a variable elimination and a stochastic sampling, efficiently solve marginal inference problems on the variational models. We review the literature of statistical relational models and variational inference (Section 8). We show that the algorithm gives a better solution than solutions with existing methods (Section 9).

## 2 Relational Hybrid Models (RHMs)[1]

A **factor** $f = (A_f, \phi_f)$ is a pair, composed of a tuple of random variables (**rvs**) $A_f$ and a potential function $\phi_f$. Here, $\phi_f$ is an unnormalized probability density from the range of $A_f$ to the nonnegative real numbers. The range of a rv can be discrete or continuous, i.e., hybrid domains. Given a **valuation v** of rvs, the **potential** of $f$ on **v** is $\phi_f(v)$.

We define **parameterized (indexed) rvs** by using **predicates** those are functions mapping parameter values to rvs. A **relational atom** (or just **atom**) denotes a parametrized rv with free parameter variable(s). For example, an atom **X(a)** can be mapped to one of n rvs $\{X(a_1), \cdots, X(a_n)\}$ when the free parameter variable **a** is **substituted** by a value $a_i$.

A **parfactor** $g = [L, A_g, \phi_g]$ is a tuple composed of a set of parameters L, a tuple of relational atoms $A_g$ and a potential function $\phi_g$. A substitution $\theta$ is an assignment to $L$, and $A_g\theta$ the relational atom (possibly ground) resulting from replacing the logical variables by their values in $\theta$. $gr(g)$ is a set of factors derived from the parfactor $g$ by substitutions.

Following example is a parfactor:

$$[ \underbrace{(\mathbf{a}, \mathbf{b})}_{\text{Parameter variables}}, \underbrace{(\mathbf{X(a)}, \mathbf{Y(b)})}_{\text{Relational atoms}}, \underbrace{f_\mathcal{N}(\mathbf{X(a)} - \mathbf{Y(b)}; \mu, \sigma^2)}_{\text{A potential (linear Gaussian)}} ]. \quad (1)$$

The domains of the parameter variables (**a** and **b**) are $\{a_1, \ldots, a_n\}$ and $\{b_1, \ldots, b_m\}$. Thus, any substitution (e.g. $\mathbf{a} = a_i, \mathbf{b} = b_j$) let two rvs (e.g. $\mathbf{X(a_i)}, \mathbf{Y(b_j)}$) hold the linear Gaussian relationship $f_\mathcal{N}$.

A **Relational Hybrid Model (RHM)** is a compact representation of graphical models with discrete and continuous rvs. An RHM is composed of a **domain**, the set of possible parameter values, and a set of parfactors **G**. The joint probability of an RHM $G$ on a valuation $v$ of rvs is as follows:

$$\frac{1}{z} \prod_{g \in \mathbf{G}} \prod_{f \in gr(g)} \phi_f(\mathbf{v})$$

where $z$ is the normalizing constant.

This representation is rather straightforward. However, inference procedures often result in complex models. For example, eliminating $X(a_1)$ in Equation (1) makes all other rvs fully connected. To address this problem, we focus on an important property of RHMs such that ground rvs mapped from a relational atom are **exchangeable**, defined as follows:

**Definition** (Exchangeable Random Variables). A sequence of rvs $X(a_1), \cdots, X(a_n)$ is **exchangeable**, when for any finite permutation $\pi()$ of the indices the joint probability of the permuted sequence $X(a_{\pi(1)}), \cdots, X(a_{\pi(n)})$ is the same as the joint probability of the original sequence.

Note that RHMs may include atoms with non-exchangeable rvs.[2] In this case, our variational algorithm grounds, or shatters, any atom including non-exchangeable rvs. That is, the atom degenerates into a set of propositional rvs. The detail conditions to determine exchangeable rvs, see [26; 21; 4; 12].

For convenience, we use $\mathbf{X}^n$ to refer to the set of n rvs, which are mapped from a relational atom $X(\mathbf{a})$ by substitutions, i.e., $\mathbf{X}^n = \{X(a_1), \cdots, X(a_n)\}$. The joint probability of the rvs mapped from two atoms $X(\mathbf{a})$ $Y(\mathbf{b})$ can be represented as follows: $P(\mathbf{X}^n, \mathbf{Y}^m) = P(X(a_1), \cdots, X(a_n), Y(b_1), \cdots, Y(b_m))$.

Potentials with a large number of rvs in RHMs introduce several difficulties in representation, learning and inference. To address these difficulties, we propose a model-factorization based on a variational method and de Finetti's theorem [11].

## 3 Background: Variational Inference

Variational methods are used to convert a complex problem into a simpler problem, where the simpler problem is charaterized by a decoupling of the degrees of freedom in the original problem [19]. As an example of *ground* models, one variational representation of model $P(x_1, \cdots, x_n)$ can be as follows:

$$P(x_1, \cdots, x_n) \approx \sum_\theta \prod_i P(x_i, |\theta) P(\theta)$$

where $\theta$ is a latent random variable. Inference is a procedure which computes marginal probabilities, say $P(X')$ ($X' \in \{x_1, \cdots, x_n\}$). This can be done by eliminating (or summing out) all variables except ones in $X'$. The variational representation (right-hand side) allows efficient inference procedures because all random variables are factorized.

**De Finetti-Hewitt-Savage's Theorem:** For exchangeable rvs, de Finetti's theorem [11] showed that any probability distribution $P(\mathbf{X}^n)$ of an infinite number $n$ of binary exchangeable rvs can be represented by a mixture of independent and identically distributed

---
[1] Parts of our model representations in this section are based on the previous works ([26; 12; 23; 6]).

[2] Lifted inference for RHMs with non-exchangeable rvs is out of scope of this paper. There are recent developments to handle some of such cases, e.g. [17; 2].

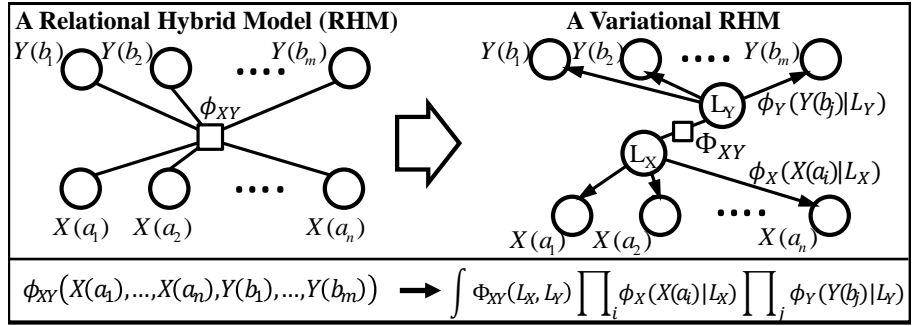

Figure 1: An illustration of factoring a potential $\phi_{XY}(\mathbf{X}^n, \mathbf{Y}^m)$. Our algorithm converts an RHM (left) into a variational (or factored) RHM (right) where the probability is represented by two latent variables $L_X$ and $L_Y$.

(**iid**) Bernoulli rvs $P_{Miid}(\mathbf{X}^n)$ with a parameter $\theta$:

$$\lim_{n\to\infty} P(\mathbf{X}^n) = \int_0^1 \theta^{t_n}(1-\theta)^{n-t_n} \Phi_X(\theta)\, d\theta = P_{Miid}(\mathbf{X}^n),$$

where $t_n = \sum_i X(a_i)$, and $0 \le \theta \le 1$ is a **latent variable**. This observation is extended to multi-valued and continuous rvs by [16].

$$\lim_{n\to\infty} P(\mathbf{X}^n) = \int \prod_{i=1}^n \phi_X(X(a_i)|L_X) \Phi_X(dL_X) = P_{Miid}(\mathbf{X}^n), \quad (2)$$

where $L_X$ is a **latent variable** which chooses a distribution $\phi_X(\mathbf{X}(a)|L_X)$ of the iid rvs.[3] We represent the variational form as $P_{Miid}(\mathbf{X}^n)$ (or $\phi_{Miid}(\mathbf{X}^n)$ for unnormalized potentials).[4] Compared to the input distribution $P(\mathbf{X}^n)$, the number of parameters of the variational form, e.g. entries of conditional density tables in $P_{Miid}(\mathbf{X}^n)$, can be substantially reduced by this factorization. As shown in Figure 1, when the variational models are applied to two sets of exchangeable rvs, the variational model (the right hand side) requires parameters of $\phi_X(X(a_i)|L_X)$, $\phi_Y(Y(b_j)|L_Y)$ and $\Phi_{XY}(L_X, L_Y)$ not parameters of $\phi_{XY}(X(a_i), \cdots, Y(b_m))$.

We present variational representations for multiple sets of finite, exchangeable rvs, and new error anaysis (Section 7). Beforehand, we introduce our variational learning and inference algorithms.

## 4 Algorithms: Lifted Variational Inference

This section outlines our pragmatic variational inference algorithm, *Lifted Relational Variational Inference* (**LRVI**). LRVI is composed of two main subroutines: learning a variational approximation, *Find-Variational-RHM*; and eliminating latent variables (inference) *Latent-Variable-Elimination* which can be replaced by *Lifted-MCMC* (Section 6.3).

---

[3] Here, we consider $\Phi_X$ with a density. Thus $\Phi_X(dL_X)$ in Equation (2) can be replaced by $\Phi(L_X)\, dL_X$.

[4] Here, *Miid* stands for a mixture of iid rvs.

---

**Input**: $G$ an *RHM* (a set of parfactors), $Q$ a query (a set of relational atoms), $O$ observations
**Output**: $P(Q)$ a (posterior) distribution of $Q$
**begin**
    // (One-time) Variational Learning
    **if** $G \notin \{Variational\ RHMs\}$ **then**
        $G \leftarrow$ **Find-Variational-RHM**$(G)$;
    // Main Inference Routine
    $P(Q) \leftarrow$ **Latent-Variable-Elimination** $(G, Q, O)$;
    **return** $P(Q)$;

**Algorithm Lifted Relational Variational Inference:** It receives an RHM, a query and observations, then returns a posterior of the query.

---

LRVI receives an RHM $G$, a query $Q$ and observations $O$ as inputs. It outputs the conditional probability, $P(Q|O)$. In the routine, it examines that each potential in $G$ is the variational form, a mixture of product of iid rvs. If not, it calls *Find-Variational-RHM(G)* and receives a variational RHM $G_{Miid}$. The variational RHM is calculated once, and reused next time. With the $G_{Miid}$, *Latent-Variable-Elimination*($G_{Miid}$, $Q$, $O$) solves the inference problem $P(Q|O)$. This is done by the variable elimination which iteratively eliminates non-query atoms.

---

**Input**: $G$, an RHM
**Output**: $G_{Miid}$, a variational RHM
**begin**
    **for** $g = (L, A, \phi) \in G$ **do**
        **if** $A$ *has no continuous atom* **then**
            $\phi_{Miid} \leftarrow$ **Lifting-Discrete**$(\phi)$ (Section 5.1);
        **else**
            $\phi_{Miid} \leftarrow$ **Lifting-Continuous**$(\phi)$ (Section 5.2);
        $G_{Miid} \leftarrow G_{Miid} \cup \{(L, A, \phi_{Miid})\}$;
    **return** $G_{Miid}$;

**Algorithm Find-Variational-RHM:** It finds a variational approximation for an input RHM (Section 5).

---

*Find-Variational-RHM(G)* converts the potential $\phi$ in each parfactor into a variational potential $\phi_{Miid}$, a mixture of iid rvs as shown in Equation (2). For potentials with only discrete atoms, it calls *Lifting-Discrete($\phi$)* and receives a variational potential $\phi_{Miid}$. For potentials including at least one continuous atom, it calls *Lifting-Continuous($\phi$)*. After iterating and converting all par-factors in $G$, a variational RHM $G_{Miid}$ is returned. Section 5 explains the procedures in detail.

---
**Input**: $G_{Miid}$ a variational *RHM*, $Q$ a query, $O$ observations
**Output**: $P(Q)$ a (posterior) distribution of $Q$
**begin**
    $G_{Miid} \leftarrow$ Update-Obs($G_{Miid}, O$) ; // For observations
    $\mathbf{A} \leftarrow$ a set of atoms in $G_{Miid}$; $\mathbf{\Phi} \leftarrow \{\phi_g | g \in G_{Miid}\}$;
    **for** $X \in \mathbf{A} \setminus Q$ **do**
        $\Phi_X \leftarrow \{\phi \in \mathbf{\Phi} | X$ is argument of $\phi\}$;
        **if** $X$ *is discrete* **then**
            $\phi' \leftarrow$ **Inference-Discrete**($\Phi_X$) (Section 6.1);
        **else**
            $\phi' \leftarrow$ **Inference-Continuous**($\Phi_X$) (Section 6.2);
        $\mathbf{\Phi} \leftarrow (\mathbf{\Phi} \setminus \phi_X) \bigcup \{\phi'\}$ ;
    **return** $\mathbf{\Phi}$;

---
**Algorithm Latent-Variable-Elimination:** It sums out non-query latent variables, and returns a posterior of the query (Section 6).

---

*Latent-Variable-Elimination($G_{Miid}, Q, O$)* call *Update-Obs($G_{Miid}, O$)* to update the potentials of latent variables based on observations $O$. The intuition of *Update-Obs* is that each rv is conditionally independent given latent variables like the Naive Bayes models [18]. Variational RHMs allows a simple update algorithm to maintain the relational structure upon individual observations.[5] It iteratively eliminates all latent variables except the query without referring to ground variables. Section 6 includes detail procedures.

## 5 Variational Learning for RHMs

This section elaborates a learning algorithm which converts each potential in an RHM into a variational potential. Here, the key procedure is to extract the potential on latent variables (e.g. $\Phi_X(L_X)$).

The potential on latent variables can be derived analytically and exactly, when an input potential satisfies some conditions such as $\infty$-extendible (explained in Section 7). It is also known that discrete potentials[6] and some Gaussian potentials (e.g. pairwise Gaussian [5] and Gaussian processes [9; 31]) allow such derivations.

Unfortunately, it is hard to use such derivations in general hybrid models because many real-world potentials are neither $\infty$-extendible nor Gaussian. Here, we present our solutions for (more intuitive) discrete models first, and then for continuous models.

### 5.1 Lifting Discrete Potentials

For discrete potentials, we need to find the probability density $\Phi_X(L_X)$ over the iid (Bernoulli) rvs where

---
[5]Existing lifted inference methods degenerate relational models upon observations. See Split [26] or Shatter [12].
[6]Section 8 includes an empirical comparison with existing lifted inference for discrete models

---

$L_X$ is the Bernoulli parameter. To represent an input potential $\phi(\mathbf{X}^n)$ compactly, we group equivalent value assignments according to the **value-histogram representation** [12; 23], $\phi(\mathbf{X}^n) = \phi_\mathbf{h}(h_X)$.[7] Learning variational parameters of Equation (2) with discrete rvs is formulated as follows:

$$\arg\max_{\Phi_X(p)} \left\| \phi(\mathbf{X}^n) - \int \Phi_X(p) \prod_{i=1}^n \phi_X(X(a_i)|p) \, dp \right\|$$
$$= \arg\max_{\Phi_X(p)} \left\| \phi_\mathbf{h}(h_X) - \int \Phi_X(p) \cdot f_{\mathcal{B}/\mathcal{M}}(h_X; n, p) \, \mathbf{d}p \right\|$$
$$\approx \arg\max_{\mathbf{w}, p_\mathbf{X}} \left\| \phi_\mathbf{h}(h_X) - \sum_{l=1}^k w_l \cdot f_{\mathcal{B}/\mathcal{M}}(h_X; n, p_{X_l}) \right\|, \quad (3)$$

where $\|P - Q\|$ (or $d_{TV}(P, Q)$) is the **total variation distance**[8]; $f_{\mathcal{B}/\mathcal{M}}(h_X; n, p)$ is a binomial (or multinomial) pdf; $\mathbf{w} = (w_1, \cdots, w_k)$ is a $k$-dimensional weight vector such that $w_l = \Phi_X(p_{X_l})$, $\sum_{l=1}^k w_l = 1$; and $p_\mathbf{X} = (p_{X_1}, \cdots, p_{X_k})$ is a vector of $k$ values chosen from $[0, 1]$, the domain of the latent variable $p$.

For **binary** exchangeable rvs, the iid potential $\phi_X(X(a_i)|p)$ is the Bernoulli distribution with a parameter $p$ (i.e. $P(X(a_i)) = p$). When equivalent models in $\mathbf{X}^n$ are grouped into the histogram $h_X$, the variational terms can be represented as a binomial distribution (the second line in Equation (3)) because $\binom{n}{h_X} \prod_i \phi_X(X(a_i)|p) = f_{\mathcal{B}}(h_X; n, p)$. That is, the problem is reduced to learn a mixture of $k$ binomial distributions where $w_l$ is a weight for each binomial $f_{\mathcal{B}}(h_X; n, p_{X_l})$.

For **multi-valued** exchangeable rvs, the iid potential $\phi_X(X(a_i)|p)$ is the Categorical distribution, i.e. multi-valued Bernoulli. Thus, this problem is reduced to learn a mixture of multinomial distributions $f_\mathcal{M}$:

$$\arg\max_{\mathbf{w}, p_\mathbf{X}} \left\| \phi_\mathbf{h}(h_X) - \sum_{l=1}^k w_l \cdot f_\mathcal{M}(h_X; n, p_{X_l}) \right\|. \quad (4)$$

For potentials with two or more atoms, it can be formulated as follows:

$$\arg\max_{\mathbf{w}, p_\mathbf{X}, p_\mathbf{Y}} \left\| \phi_\mathbf{h}(h_X, h_Y) - \sum_{l=1}^k w_l f_{\mathcal{B}/\mathcal{M}}(h_X; n, p_{X_l}) f_{\mathcal{B}/\mathcal{M}}(h_Y; m, p_{Y_l}) \right\|,$$

where $p_\mathbf{Y}$ is a k-dimensional vector, $(p_{Y_1}, \cdots, p_{Y_k})$; and $f$ is either the binomial or the multinomial depending on the domain of rvs.

We learn a mixture of binomials (or multinomials), e.g. $(\mathbf{w}, p_\mathbf{X})$ in Equation (3), from the original potential $\phi(\mathbf{X}^n)$ using an incremental EM algorithm.[9] Because the $k$ is not known or given, the incremental EM algorithm increases $k$ up to $n$ until the variational error converges. Assuming that the EM algorithm interates up to a constant times given a fixed k, the computational complexity of the incremental EM algorithm for $n$ binary, exchangeable rvs is bounded by $O(n^3)(= \sum_{k=1}^{n} c \cdot O(kn))$. Note that, the algorithm increases $k$ from 1 to $n$. In each EM step, $k$ components visit $n$ histogram entries $O(kn)$.

---
[7]$h_X$ is a vector with $h_{X_v} = |\{i : X(a_i) = v\}|$.
[8]$\|P-Q\| = \sup_{A \in \mathcal{B}}(P(A) - Q(A))$ when $\mathcal{B}$ is a class of Borel sets.
[9]EM algorithms are common to learn parameters for mixture models [30; 10; 27].

### 5.2 Lifting Continuous and Hybrid Potentials

For a potential $\phi(\mathbf{X}^n)$ with continuous rvs, we use a mixture of non-parametric densities to represent variational potentials. Here, we generate samples from the input potential, then learn parameters for the mixture of non-parametric densities.

Equation (2) is used to formulate the learning problem as follows:

$$\underset{\Phi_X(L_X)}{\arg\max} \left\| \phi(\mathbf{X}^n) - \int \Phi_X(L_X) \cdot \prod_{i=1}^{n} \hat{f}_{L_X}(X(a_i)) \, \mathbf{d}L_X \right\| \quad (5)$$

$$\approx \underset{\mathbf{w}, \hat{f}_\mathbf{X}}{\arg\max} \left\| \phi(\mathbf{X}^n) - \sum_{u=1}^{k} w_u \cdot \prod_{i=1}^{n} \hat{f}_{X_l}(X(a_i)) \right\|, \quad (6)$$

where $\hat{f}_{L_X}$ and $\hat{f}_{X_l}$ refer to (non-parametric) probability distributions. To solve the optimization problem, we generate $N$ samples $\mathbf{v_1}, \cdots, \mathbf{v_N}$ from the input potential $\phi(\mathbf{X}^n)$ where $\mathbf{v_t} = (v_{t_1}, \cdots, v_{t_n})$ is a $n$-dimensional vector, value assignments for $n$ rvs. Then, we solve the following maximum likelihood estimation (MLE) problem: $\arg\max_{\mathbf{w}, \hat{f}_\mathbf{X}} \sum_{t=1}^{N} \log \left( \sum_{l=1}^{k} w_l \cdot \prod_{i=1}^{n} \hat{f}_{X_l}(v_{t_i}) \right)$, where we denote the kernel density estimator by $\hat{f}_{X_l}(x) = \frac{1}{S\sigma^2} \sum_{i=1}^{S} K\left(\frac{x-\mu_i}{\sigma^2}\right)$ where $(\mu_1, \cdots, \mu_S)$ are $S$ data points that underlie the density, and $\sigma^2$ is a parameter. For simplicity, we use the Gaussian Kernel, $K(x) = \frac{1}{\sqrt{2\pi}} e^{-x^2/2}$.

It is interesting to note that the kernel density estimator is analogous to the value-histogram for discrete rvs [12; 23] in a sense that frequently observed regions (or bins) have the higher probability. This new insight enables us to represent continuous models compactly.

For potentials with two or more atoms, the approach can be formulated as follows:

$$\underset{\mathbf{w}, \hat{f}_\mathbf{X}, \hat{f}_\mathbf{Y}}{\arg\max} \sum_{t=1}^{N} \log \left( \sum_{l=1}^{k} w_l \cdot \prod_{i=1}^{n} \hat{f}_{X_l}(v_{t_i}^X) \cdot \prod_{j=1}^{m} \hat{f}_{Y_l}(v_{t_j}^Y) \right),$$

where $v_t^X$ and $v_t^Y$ are respectively the $t^{\text{th}}$ samples of $\mathbf{X}^n$ and $\mathbf{Y}^m$.

This MLE problem is also solved by an EM algorithm. $N$ samples are used to build $k$ densities in the maximization (M) step, and the likelihood of each sample is calculated in the expectation (E) step. variation error converges.

## 6 Lifted Inference with RHMs

In this section we build on the result of previous sections and present lifted inference algorithms that utilize the learned variational models to speed up relational inference. The lifted inference algorithms find solutions without referring to ground rvs.

*Latent-Variable-Elimination* marginalizes relational atoms with the following steps: (i) choosing an atom; (ii) finding all potentials including the atom and making the product of them; (iii) summing out the atom; and (iv) repeating the steps until only query atoms are left. We demonstrate the key step (iii) with two variational potentials, $\phi_{Miid}(\mathbf{X}^n, \mathbf{Y}^m)$ and $\phi'_{Miid}(\mathbf{Y}^m)$.

### 6.1 Inference with Discrete Variables

The key intuition is that the variational form is maintained after eliminating an atom. We demonstrate the intuition by an example. The marginal probability of the latent variable $L_X$ is calculated by eliminating (or summing) $\mathbf{Y}^m$ out: $\sum_{h_y} \phi_\mathbf{h}(h_x, h_y) \cdot \phi'_\mathbf{h}(h_y)$

$$\approx \sum_{h_y} \sum_{l=1}^{k} w_l f_\mathcal{B}(h_x; n, p_{X_l}) f_\mathcal{B}(h_y; m, p_{Y_l}) \sum_{l'=1}^{k'} w'_{l'} f_\mathcal{B}(h_y; m, p_{Y_{l'}})$$

$$= \sum_{l=1}^{k} \sum_{l'=1}^{k'} w_l w'_{l'} \left( \sum_{h_y} f_\mathcal{B}(h_y; m, p_{Y_l}) f_\mathcal{B}(h_y; m, p_{Y_{l'}}) \right) f_\mathcal{B}(h_x; n, p_{X_l})$$

$$\approx \sum_{l=1}^{k} \sum_{l'=1}^{k'} w_l w'_{l'} \left( \int f_\mathcal{N}(h_y; \mu_l, \sigma_l^2) f_\mathcal{N}(h_y; \mu_{l'}, \sigma_{l'}^2) \, \mathbf{d}h_y \right) f_\mathcal{B}(h_x; n, p_{X_l})$$

$$= \sum_{l=1}^{k} w_{\bar{Y},l} \cdot f_\mathcal{B}(h_x; n, p_{X_l}) = \phi''(\mathbf{X}^n) \quad (7)$$

when $\sum_{l=1}^{k} w_{\bar{Y},l} = 1$ and $f_\mathcal{N}(h_y; \mu_l, \sigma_l^2)$ is the Normal approximation to binomial such that $\mu_l(=m \cdot p_{Y_l})$ and $\sigma_l^2 (=m \cdot p_{Y_l} \cdot (1-p_{Y_l}))$. It is important to note that binomial pdfs are not closed under the product operation. That is, a product of two binomial pdfs are not a binomial pdf unless the binomial parameters $p_{X_l}$ $p_{X_l}$ are identical. For large $n$ and $m$, the Normal approximation to Binomial is an important step to maintain the variational structure during the inference procedure. In this way, after eliminating $\mathbf{Y}^m$, the marginal potential $\phi''_{Miid}(\mathbf{X}^n)$ is still represented as the variational form. The same principle is applied for potentials with more than two atoms.

Now, we will show that the product of variational forms in Step (iii) can also be represented as a variational form. Consider the following two variational potentials $\phi_{Miid}(\mathbf{X}^n)$ $\phi'_{Miid}(\mathbf{X}^n)$. The product operation is common during the elimination step as shown in Equation (7). The product of the two potentials is represented as follows:

$$\left( \sum_{l=1}^{k} w_l \cdot f_\mathcal{B}(h_x; n, p_{X_l}) \right) \cdot \left( \sum_{l'=1}^{k'} w'_{l'} \cdot f'_\mathcal{B}(h_x; n, p_{X_{l'}}) \right)$$

$$\approx \sum_{l=1}^{k}\sum_{l'=1}^{k'} w_l \cdot w_{l'}' \int f_{\mathcal{N}}(h_x;\mu_l,\sigma_l^2)\cdot f_{\mathcal{N}}'(h_x;\mu_{l'},\sigma_{l'}^2)\,\mathbf{d}h_x$$

$$= \sum_{l=1}^{k}\sum_{l'=1}^{k'} w_l \cdot w_{l'}' \cdot z_{l,l'} f_{\mathcal{N}}(h_x;\mu_{new},\sigma_{new}^2) = \phi_{Miid}'''(\mathbf{X}^n), \quad (8)$$

$z_{l,l'}$ is the inverse of the normalizing constant. This derivation shows that a product of variational potentials results in a variational potential as $\phi_{Miid}'''(\mathbf{X}^n)$.[10]

### 6.2 Inference with Continuous Variables

For continuous variables, we also demonstrate the intuition by an example with two potentials $\phi_{Miid}(\mathbf{X}^n,\mathbf{Y}^m)$ $\phi_{Miid}'(\mathbf{Y}^m)$ where $\mathbf{X}^n$ and $\mathbf{Y}^m$ are two sets of continuous rvs. Each potential is represented as shown in Section 5.2. When we eliminate $\mathbf{Y}^m$, it can be formulated as follows:

$$\int \left( \sum_{l=1}^{k} w_l \prod_{i=1}^{n} \hat{f}_{X_l}(X(a_i)) \prod_{j=1}^{m} \hat{f}_{Y_l}(Y(b_j)) \right) \phi_{Miid}'(\mathbf{Y}^m)\,\mathbf{d}\mathbf{Y}$$

$$= \sum_{l=1}^{k}\sum_{l'=1}^{k'} w_l w_{l'} \prod_{i=1}^{n} \hat{f}_{X_l}(X(a_i)) \prod_{j=1}^{m} \left( \int \hat{f}_{Y_l}(Y(b_j)) \hat{f}_{Y_{l'}}(Y(b_j))\,\mathbf{d}Y(b_j) \right)$$

$$= \sum_{l=1}^{k}\sum_{l'=1}^{k'} w_l w_{l'} \frac{1}{z_{l,l'}{}^m} \prod_{i=1}^{n} \hat{f}_{X_l}(X(a_i)) = \phi_{Miid}''(\mathbf{X}^n), \quad (9)$$

$z_{l,l'}$ is the normalizing constant calculated from the product of two mixtures of Normals: $\hat{f}_{Y_l}(Y(b_j))$; and $\hat{f}_{Y_{l'}}(Y(b_j))$.

Finally, we show that the product of two variational potentials becomes a variational form: $\left(\sum_{l=1}^{k} w_l \cdot \prod_{i=1}^{n} \hat{f}_{X_l}(X(a_i))\right) \cdot \left(\sum_{l'=1}^{k'} w_{l'}' \cdot \prod_{i=1}^{n} \hat{f}_{X_{l'}}(X(a_i))\right)$

$$= \sum_{l=1}^{k}\sum_{l'=1}^{k'} w_l w_{l'}' \prod_{i=1}^{n} \hat{f}_{X_l}(X(a_i)) \hat{f}_{X_{l'}}(X(a_i))$$

$$= \sum_{l=1}^{k}\sum_{l'=1}^{k'} w_l w_{l'}' \frac{1}{z_{l,l'}{}^n} \cdot \prod_{i=1}^{n} \hat{f}_{X_{l,l'}}^{new}(X(a_i)) = \phi_{Miid}'''(\mathbf{X}^n). \quad (10)$$

### 6.3 Lifted Markov chain Monte Carlo (MCMC)

When variational RHMs include a large number of mixture compomcnts. The latent variable elimination may take long time. In this case, we use a lifted MCMC algorithm: (i) choosing a latent variable (e.g $L_X$) randomly; (ii) calculating the conditional probability of the latent variable (e.g. $\Phi_X(L_X)$) using assignment of neighboring latent variables; (iii) choosing an assignment from the probability (e.g. $L_X = p_{X_l}$ ($1 \le l \le k$)); and (iv) repeating until convergence.

---

[10]$\phi_{Miid}'''(\mathbf{X}^n)$ is a mixture of $|k \cdot k'|$ Normals. When $|k \cdot k'|$ is large, it is possible to collapse the mixture into a mixture of fewer components.

---

Here, the steps (ii) and (iii) are main steps. Step (ii) is a subset of the procedure in Equations (7) and (9), because the values of neighboring latent variables (e.g. $L_Y = p_{Y_{l'}}$ ($1 \le l' \le k'$)) can be simply assigned. Step (iii) is a straightforward sampling procedure which chooses one component based on its weight. For example, $w_l \cdot w_{l'}' \cdot z_{l,l'}$ in Equation (8) is a weight for one of $|k| \cdot |k'|$ Normal distributions in $\phi_{Miid}'''(\mathbf{X}^n)$.

## 7 Relational-Variational Lemmas

This section provides error analysis of our variational approximations for a single atom, multiple atoms and the general case (RHMs). Beforehand, we define a term, $\overline{\mathbf{n}}$-extendible:

**Definition** ($\overline{\mathbf{n}}$-extendible). $P(\mathbf{X}^n)$, a probability with $n$ exchangeable rvs, is $\overline{\mathbf{n}}$-extendible when the followings hold: (1) there is $P(\mathbf{X}^{\overline{n}})$, a probability with $\overline{n}$ exchangeable rvs ($\overline{n} > n$); and (2) $P(\mathbf{X}^n)$ is the marginal distribution of $P(\mathbf{X}^{\overline{n}})$, i.e., eliminating ($\overline{n} - n$) rvs.

Figure 2 explains the intuition of $\overline{\mathbf{n}}$-extendible potentials for discrete models. If a potential is not extendible, it has rugged bars, e.g. a single peak. If a potential is extendible to a large number $\overline{n}$, the potential has a smoothed histogram. If a potential is ∞-extendible, it is represented by a mixture of binomials exactly.

**Lemma 1.** [13] *If $P(\mathbf{X}^n)$, a probability with n exchangeable rvs, is $\overline{n}$-extendible, then the **total variation distance** $d_{TV}(P(\mathbf{X}^n), P_{Miid}(\mathbf{X}^n))$ of the input probability $P(\mathbf{X}^n)$ and the variational form $P_{Miid}(\mathbf{X}^n)$ in Equation (2) is bounded as follows: (i) when $X(a_i)$ are d-valued discrete rvs, $d_{TV}(P(\mathbf{X}^n), P_{Miid}(\mathbf{X}^n)) \le \frac{2dn}{\overline{n}}$; and (ii) when $X(a_i)$ are continuous rvs, $d_{TV}(P(\mathbf{X}^n), P_{Miid}(\mathbf{X}^n)) \le \frac{n(n-1)}{\overline{n}}$.*

### 7.1 Our Results: Variational RHMs

**Factoring Potentials with Multiple Atoms:** De Finetti-Hewewitt-Savage's theorem (Section 3) is applicable only to potentials with a single atom. Here, we present our new theoretical results on variational RHMs.

**Lemma 2. [Existence of a Variational Form]** *For $P(\mathbf{X}^n, \mathbf{Y}^m)$, a probability with two atoms in an RHM, there are two new latent variables, $L_X$ and $L_Y$, and a new potential $\Phi_{XY}(L_X, L_Y)$ such that the following holds,* $\lim_{n,m \to \infty} P(\mathbf{X}^n, \mathbf{Y}^m)$

$$= \int \Phi(L_X, L_Y) \prod_{i=1}^{n} \phi_X(X(a_i)|L_X) \prod_{j=1}^{m} \phi_Y(Y(b_j)|L_Y)\,\mathbf{d}L_X L_Y$$

$$= P_{Miid}(\mathbf{X}^n, \mathbf{Y}^m).$$

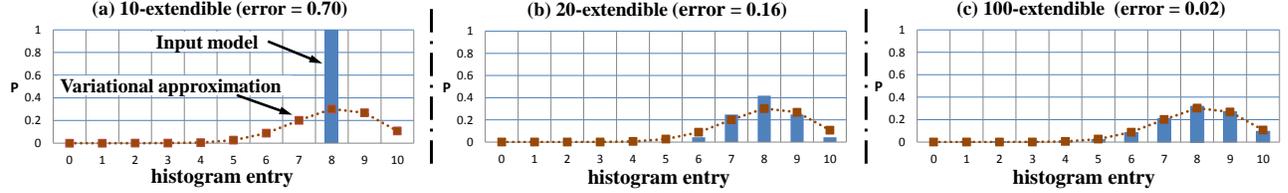

Figure 2: Illustrations of three different value-histograms of 10 exchangeable binary rvs. Dotted lines with markers represent the best possible variational approximation, i.e., a mixture of binomials. (a), (b) and (c) respectively present potentials extendible up to 10, 20 and 100 rvs. For the potential in (1), the variational approximation has a high error, total variation, and thus is a poor approximation. When a potential is extendible to the larger number, the variational approximation is smaller as shown in (2) and (3).

*Sketch of proof.* Assigning values to one atom is fixed (e.g. $\mathbf{Y}^m = \mathbf{v}$) results in a new potential with one atom, $\phi(\mathbf{X}^n)$, which can be factored into $\int \Phi(L_X|\mathbf{v}) \prod_i \phi_X(X(a_i)|L_X) \mathbf{d}L_X$. Because $\Phi(L_X|\mathbf{v})$ is conditioned on $\mathbf{v}$, $\Phi(L_X|\mathbf{Y}^m)$ can be factored into $\int \Phi(L_X, L_Y) \prod_j \phi_Y(Y(b_j)|L_Y) \mathbf{d}L_Y$. □

To analyze variational error of potentials with multiple atoms, we introduce another term $(\overline{\mathbf{n}}, \overline{\mathbf{m}})$-**extendible**.

**Definition** (($\overline{\mathbf{n}}, \overline{\mathbf{m}}$)-extendible). $P(\mathbf{X}^n, \mathbf{Y}^m)$ is $(\overline{\mathbf{n}}, \overline{\mathbf{m}})$-**extendible** when (1) there is $P(\mathbf{X}^{\overline{n}}, \mathbf{Y}^{\overline{m}})$, a probability with two sets of exchangeable rvs ($\overline{n} > n, \overline{m} > m$); and (2) $P(\mathbf{X}^n, \mathbf{Y}^m)$ is the marginal distribution of $P(\mathbf{X}^{\overline{n}}, \mathbf{Y}^{\overline{m}})$.

**Lemma 3** (*Error of the Variational Parfactor*). *If $P(\mathbf{X}^n, \mathbf{Y}^m)$, a probability with two exchangeable rvs in an RHM, is $(\overline{n}, \overline{m})$-extendible, then the total variation $d_{TV}(P(\mathbf{X}^n, \mathbf{Y}^m), P_{Miid}(\mathbf{X}^n, \mathbf{Y}^m))$ is bounded as follows: (i) when $\mathbf{X}^n$ and $\mathbf{Y}^m$ are respectively $d_x$-valued and $d_y$-valued discrete rvs, $d_{TV}(P, P_{Miid}) \leq \frac{2d_x n}{\overline{n}} + \frac{2d_y m}{\overline{m}}$; (ii) when $\mathbf{X}^n$ are $d_x$-valued discrete and $\mathbf{Y}^m$ are continuous, $d_{TV}(P, P_{Miid}) \leq \frac{2d_x n}{\overline{n}} + \frac{m(m-1)}{\overline{m}}$; (iii) when $\mathbf{X}^n$ and $\mathbf{Y}^m$ are continuous, $d_{TV}(P, P_{Miid}) \leq \frac{n(n-1)}{\overline{n}} + \frac{m(m-1)}{\overline{m}}$.*

*proof.* The intuition is that the error of a variational model is additive with an additional atom. We used the principle in [13] such that a $\bar{n}$-extendible pdf, $P(X^n)$, can be represented by a mixture of extreme pdfs (e.g. $P(X^n) = \sum_e w_e p_e$). Here, an extreme pdf $p_e$ is a probability of $n$ draws made at random without replacement from an urn, $U$, which contains $\bar{n}$ balls marked by one of $c$ colors. Let $e$ a unique marking in $U$. The variation distance of each extreme point $p_e$ and its variational form $P_{Miid}(X^n|e)$ is bounded $\leq \frac{2cn}{\bar{n}}$ for discrete rvs.

For a distribution with the multiple atoms, each extreme point corresponds to the joint distribution of $n$ draws from $U_X$ of $\bar{n}$ balls, and $m$ draws from $U_Y$ of $\bar{m}$ balls. Because the draws can be done independently for each urn, an extreme pdf (e.g. $p_{e_x,e_y}$) can be represented as the product of independent extreme pdfs (e.g. $p_{e_x} \cdot p_{e_y}$). From the Lemma 1, the variation distances of variational forms of $p_{e_x}$ and $p_{e_y}$ are respectively bounded. WLOG, we can represent the errors with $\epsilon_x$ and $\epsilon_y$,

$$d_{TV}(p_{e_x}, \phi_{iidx}) \leq \epsilon_x, d_{TV}(p_{e_y}, \phi_{iidy}) \leq \epsilon_y.$$

Note that,

$$\begin{aligned}
&d_{TV}(p_{e_x} p_{e_y}, \phi_{iidx} \phi_{iidy}) \\
&\leq d_{TV}(p_{e_x} p_{e_y}, p_{e_x} \phi_{iidy}) + d_{TV}(\phi_{iidy} p_{e_x}, \phi_{iidy} \phi_{iidx}) * \\
&\leq d_{TV}(p_{e_y}, \phi_{iidy}) + d_{TV}(p_{e_x}, \phi_{iidx}) ** = \epsilon_x + \epsilon_y.
\end{aligned}$$

The second step (marked as ∗) is derived from the following equations (Here, $A$ is $p_{e_x} \phi_{iidy}$):

$$\begin{aligned}
p_{e_x} p_{e_y} - \phi_{iidx} \phi_{iidy} &= p_{e_x} p_{e_y} - \phi_{iidx} \phi_{iidy} - A + A \\
&= (p_{e_x} p_{e_y} - A) + (A - \phi_{iidx} \phi_{iidy}).
\end{aligned}$$

The third step (marked as ∗∗) is derived from the fact that $p_{e_x}$ and $\phi_{iidy}$ are pdfs, e.g. $\sum_{X^n} p_{e_x}(X^n) = 1$. That is, $d_{TV}(p_{e_x} p_{e_y}, p_{e_x} \phi_{iidy}) \leq d_{TV}(p_{e_y}, \phi_{iidy})$. The derivation can be applied to $\phi_{iidy}$ and continuous cases.

Thus, the total variation distance between two probabilities, $P(X^n, Y^m)$ and $P_{Miid}(X^n, Y^m)$ is bounded by the sum of two error bounds. □

Now, we present error analsys for RHMs with more than two atoms, e.g. $P(\mathbf{X}^n, \mathbf{Y}^m, \mathbf{Z}^u)$.

**Theorem 4** (*Error of Variational RHMs*). *Let $\mathbf{X}_g$ and $\mathbf{X}_G$ is respectively the set of all rvs in a parfactor $g$ and an RHM $G$. Let $P(\mathbf{X}_g)(= \frac{1}{z_g} \prod_{f \in gr(g)} w_f(\mathbf{X}_f))$; and $P(\mathbf{X}_G)(= \frac{1}{z} \prod_{g \in G} P(\mathbf{X}_g))$. The total variation $d_{TV}(P(\mathbf{X}_G) - P_{Miid}(\mathbf{X}_G))$ is bounded by $\frac{1}{z} \sum_{g \in G} \epsilon_g$ where $\epsilon_g = d_{TV}(P(\mathbf{X}_g) - P_{Miid}(\mathbf{X}_g))$ and $z$ is the normalizing constant.*

*Sketch of proof.* We can build a fully joint probability of all relational atoms with the RHM. Then, Lemma 3 can be used to prove the total variation of the variational RHM. □

## 8 Related Work

Nonparametric Bayesian Logic (NP-BLOG) [4] presents a new variational representation for relational discrete models using the Dirichlet Process. In principle, the NP-BLOG and the variational RHMs have in common: compact representations for exchangeable rvs. The difference is that NP-BLOG handles $\infty$-extendible exchangeable, discrete rvs. Here, we investigated further for several new directions: model learning, continuous domains, and approximation errors. Our error bounds provide an in-depth understanding for models with finite exchangeable discrete rvs.

For discrete models, the value-histogram [12; 23] represents potentials with polynomial numbers of histogram entries. When aggregate operators are given, [6] shows that the histgram representations can be approximately replaced by a Normal with linear constraints. We generalize and expand the concept to compress general-purpose histogram representations.

For continuous potentials, unfortunately, the histogram representation is not applicable because it is not clear how to discretize continuous domains to build up such histograms. Thus, most existing lifted inference for continuous models are limited to Gaussian potentials [5; 7; 1]. Thus, our representation is a unique lifted inference for non-Gaussian continuous potentials.

Lifted Belief Propagation (LBP) [29] solves inference problems in many practial models by grouping rvs. Here, rvs in an atom send the same messages to neighboring rvs, and are not constrained among rvs in an atom. Instead, our lifted MCMC sends a distribution, has more expressive power, and requires fewer samples until the convergence. The advance promises that our variational models and the lifted MCMC method can be a good complement to existing sampling methods such as LBP.

## 9 Experimental Results

We provide experimental results regarding the variational approximations and the efficiency and the accuracy of our LRVI in a real-world groundwater model.

First, we analyze variational approximations on the competing workshops models C-FOVE [23] which includes two parfactors, $\phi_1(attends(X), hot(W))$ and $\phi_2(attends(X), series)$. We can exactly represent $\phi_2$ with a Binomial:

$$\phi_2(\#X[attends(X)], series)$$
$$= w(series) f_\mathcal{B}(\#X[attends(X)]; n, p(series)),$$

Here, $n$ is the number of people; $w$ and $l$ are functions: $\{\top, \bot\} \to \mathbf{R}$; and $\#X[attends(X)]$ is a histogram representation in [23] such that $|\{i|attends(X_i) = \top\}|$. $\phi_1$ is also represented by a mixture of $|W|$. $\phi_1$ is the bivariate multiplicative binomial [20] and conditionally binomial. E.g., when $\#W[hot(W)]$ is fixed to 5, $\phi_1(\#X[attends(X)], \#W[hot(W)] = 5) = w_5 f_\mathcal{B}(\#X[attends(X)]; n, l_5)$ .

When using the same parameters in [23], $\phi_1(attends(X), hot(W))$ is represented by a single binomial distribution accurately because weights of other $|W-1|$ binomials are small with -9 orders of magnitude. We make similar observations with different parameters. When we randomly choose parameters (50 times), a single binomial was enough to represent $\phi_1$ with a small total variation (< 0.001) for more than 90% (46 times). For other parameters, at most three binomials can represent $\phi_1$ with a very small error (<0.0001).

Second, we apply our variational learning algorithm to a real-world groundwater dataset shown in Figure 3 (a) Republican River Compact Administration (RRCA). The dataset is composed of measurements (water levels) at over 10,000 wells and baseflow observations at 65 gages from 1918 until 2007.[11] After calibration, the training dataset is a set of partial observations in a 480 (months) by 3420 (wells) matrix. First, we cluster the 3420 wells by k-means into 10 groups based on means and variances (approximately exchangeable).[12] From the dataset, the EM algorithm directly learns a variational model until the log-likelihood converges. As a result, from 6 to 14 mixtures of Gaussians (MoGs) are learned for each cluster. Figure 3 (b) shows some learned empirical distributions, cdfs of MoGs, with high weights from two clustered area, A and B. To represent the joint distribution over the clusters, we convert the 480X3240 input matrix into a 480 (months) by 92 (MoGs) matrix.

For each test month, we compute the empirical distribution of the query variables given the partial observations. Our lifted VE returns queries in average 0.3 secs and a ground VE inference returns in average 37.9 secs. In fact, ground and variational inference algorithms use different sizes of matrices 480X3420 and 480X92, respectively. That explains the reason why the variational inference is efficient. The average total variations are 0.35 (ground) and 0.29 (variational), where smaller is more accurate.

Third, we compare the accuracy and the efficiency of our lifted MCMC algorithm with a vanilla (ground) MCMC algorithm on an already factored variational model. The model is composed of two relational atoms: a binary atom $\mathbf{Job}^n$, saying whether each individual has a job, and a continuous atom $\mathbf{HP}^m$, saying the price change of each house. $p_{Job}$ is a latent variable, which represents the Bernoulli parameters of $\phi_{Miid}(\mathbf{Job}^n)$. $p_D$ is a latent variable (probability of market down), which represents the mixture of two Gaussians, $\phi_{Miid}(\mathbf{HP}^m) =$

$$p_D \prod_{j=1}^m f_\mathcal{N}(HP(h_j); -0.3, \sigma_D^2) + (1-p_D) \prod_{j=1}^m f_\mathcal{N}(HP(h_j); 0.1, \sigma_{UP}^2).$$

---

[11]Head predictions are available via the RRCA official website, http://www.republicanrivercompact.org.

[12]Although the means and variances do not guarantee the exchangeability in the clusters, here we focus on measuring the computational efficiency of our variational method.

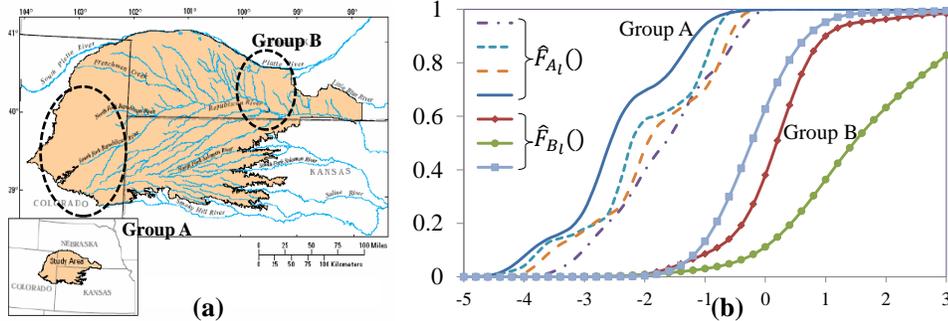

Figure 3: (a) locations of two clustered wells in the RRCA map; (b) learned empirical distributions, cdfs ($\hat{F}_{A_l}()$ and $\hat{F}_{B_l}(x)$) for two regions A and B.

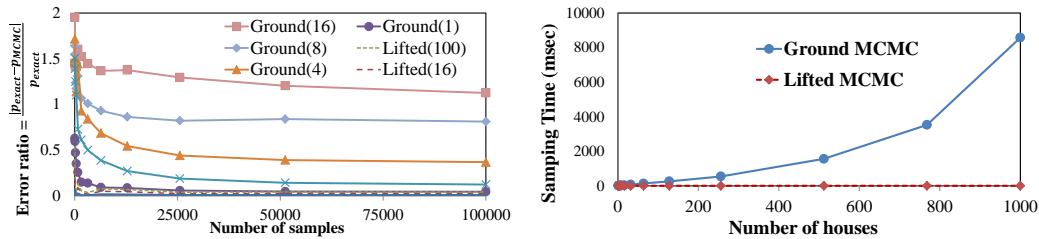

Figure 4: (a) The accuracy of the lifted MCMC and the ground MCMC with various numbers of houses. Here, () indicates the number of houses, e.g., Ground(16) is the ground MCMC with 16 houses. (b) The average time to generate samples in each MCMC step.

Here, we assume that the two latent variables follows a linear Gaussian: $\Phi(p_{Job}, p_D) = f_\mathcal{N}(p_{Job} - p_D; 0, \sigma_{JobHouse}^2)$. Figure 4 (a) shows the accuracy of the two algorithms after generating the same number of samples. That is, it measure the ratio of error to estimate a probability of a randomly chosen variable $HP(h_j)$, $|p_{exact}(HP(h_j)) - p_{MCMC}(HP(h_j))|/p_{exact}(HP(h_j))$. It shows that our lifted MCMC converges to the true density much faster than the ground MCMC.[13] Figure 4 (b) shows the average sampling time (per step) with different number of rvs, i.e. the number of houses.

## 10 Conclusions

We propose new lifted relational variational inference algorithms for relational hybrid models. Our main contributions are two folds: (1) in theory, we show that a relational model, which can represent large-scale systems, is accurately represented by a variational relational model; (2) our lifted algorithms are the first to solve inference problems without referring ground rvs for non-Gaussian continuous models. Experiments show that our method outperforms the existing possible methods in various cases including a real-world environmental problem.
## Acknowledgments

We wish to thank Wen Pu, Jihye Seong and anonymous reviewers for their valuable, constructive comments. This material is supported by NSF award ECS-09-43627 - Improving Prediction of Subsurface Flow and Transport through Exploratory Data Analysis and Complementary Modeling and DARPA award AF Sub SRI 27-001337.

---

[13] The lifted Belief Propagation (LBP) [29] is not directly applicable because the target distribution is a mixture of two iid Gaussians. The LBP assumes that each house sends the same message, i.e. price estimation. However, here, the messages among houses should be constrained. That is, the prices of some houses go down, and then prices of the other houses should go up.